\documentclass[twoside,11pt]{article}

\usepackage{pythonhighlight}
\usepackage{times}
\usepackage{balance}
\usepackage{setspace}
\usepackage{relsize}
\usepackage{verbatim} 
\usepackage{amsmath}
\usepackage{times}
\usepackage{gensymb}
\usepackage{dsfont}
\usepackage{multirow}
\usepackage{float}
\usepackage{amsfonts}
\usepackage{epstopdf}
\usepackage{mathrsfs}
\usepackage[T1]{fontenc}
\usepackage{url}
\usepackage{csquotes}
\usepackage{listings}
\usepackage{booktabs}
\usepackage{enumitem}

\usepackage{jmlr2e}
\usepackage{color}

\definecolor{codegreen}{rgb}{0,0.6,0}
\definecolor{codegray}{rgb}{0.5,0.5,0.5}
\definecolor{codepurple}{rgb}{0.58,0,0.82}
\definecolor{backcolour}{rgb}{0.95,0.95,0.92}

\lstdefinestyle{mystyle}{
	language=python,
	showtabs=true,
	tab=,
	tabsize=2,
	basicstyle=\ttfamily\footnotesize,
	stringstyle=\color{stringcolour},
	showstringspaces=false,
	alsoletter={1234567890},
	otherkeywords={\%, \}, \{, \&, \|},
	keywordstyle=\color{keywordcolour}\bfseries,
	emph={and,break,class,continue,def,yield,del,elif ,else,%
		except,exec,finally,for,from,global,if,import,in,%
		lambda,not,or,pass,print,raise,return,try,while,assert,with},
	emphstyle=\color{blue}\bfseries,
	emph={[2]True, False, None},
	emphstyle=[2]\color{keywordcolour},
	emph={[3]object,type,isinstance,copy,deepcopy,zip,enumerate,reversed,list,set,len,dict,tuple,xrange,append,execfile,real,imag,reduce,str,repr},
	emphstyle=[3]\color{commandcolour},
	emph={Exception,NameError,IndexError,SyntaxError,TypeError,ValueError,OverflowError,ZeroDivisionError},
	emphstyle=\color{exceptioncolour}\bfseries,
	morecomment=[s]{"""}{"""},
	commentstyle=\color{commentcolour}\slshape,
	emph={[4]ode, fsolve, sqrt, exp, sin, cos,arctan, arctan2, arccos, pi,  array, norm, solve, dot, arange, isscalar, max, sum, flatten, shape, reshape, find, any, all, abs, plot, linspace, legend, quad, polyval,polyfit, hstack, concatenate,vstack,column_stack,empty,zeros,ones,rand,vander,grid,pcolor,eig,eigs,eigvals,svd,qr,tan,det,logspace,roll,min,mean,cumsum,cumprod,diff,vectorize,lstsq,cla,eye,xlabel,ylabel,squeeze},
	emphstyle=[4]\color{numpycolour},
	emph={[5]__init__,__add__,__mul__,__div__,__sub__,__call__,__getitem__,__setitem__,__eq__,__ne__,__nonzero__,__rmul__,__radd__,__repr__,__str__,__get__,__truediv__,__pow__,__name__,__future__,__all__},
	emphstyle=[5]\color{specmethodcolour},
	emph={[6]assert,yield},
	emphstyle=[6]\color{keywordcolour}\bfseries,
	emph={[7]range},
	emphstyle={[7]\color{keywordcolour}\bfseries},
	literate=*%
	{:}{{\literatecolour:}}{1}%
	{=}{{\literatecolour=}}{1}%
	{-}{{\literatecolour-}}{1}%
	{+}{{\literatecolour+}}{1}%
	{*}{{\literatecolour*}}{1}%
	{**}{{\literatecolour{**}}}2%
	{/}{{\literatecolour/}}{1}%
	{//}{{\literatecolour{//}}}2%
	{!}{{\literatecolour!}}{1}%
	{[}{{\literatecolour[}}{1}%
	{]}{{\literatecolour]}}{1}%
	{<}{{\literatecolour<}}{1}%
	{>}{{\literatecolour>}}{1}%
	{>>>}{\pythonprompt}{3}%
	,%
	frame=trbl,
	rulecolor=\color{black!40},
	backgroundcolor=\color{white},
	breakindent=.5\textwidth,frame=single,breaklines=true%
}

\ShortHeadings{DESlib: A Dynamic ensemble selection library in Python}{Cruz et al.}
\firstpageno{1}
\begin{document}

\title{DESlib: A Dynamic ensemble selection library in Python}

\author{\name Rafael M. O. Cruz \email rafaelmenelau@gmail.com \\
		\name Luiz G. Hafemann \email luiz.gh@gmail.com \\
		\name Robert Sabourin \email robert.sabourin@etsmtl.ca \\
       \addr Laboratoire d'imagerie, vision et intelligence artificielle (LIVIA)\\
       \'{E}cole de Technologie Sup\'{e}rieure (\'{E}TS) - Universit\'{e} du Qu\'{e}bec\\
       Montreal, Canada
       \AND
       \name George D. C. Cavalcanti \email gdcc@cin.ufpe.br \\
       \addr Centro de Inform\'{a}tica - Universidade Federal de Pernambuco\\
       Recife, Brazil}
    
\editor{-}

\maketitle

\begin{abstract}
	
DESlib is an open-source python library providing the implementation of several dynamic selection techniques. The library is divided into three modules: (i) \emph{dcs}, containing the implementation of dynamic classifier selection methods (DCS); (ii) \emph{des}, containing the implementation of dynamic ensemble selection methods (DES); (iii) \emph{static}, with the implementation of static ensemble techniques. The library is fully documented (documentation available online on Read the Docs), has a high test coverage (codecov.io) and is part of the scikit-learn-contrib supported projects. Documentation, code and examples can be found on its GitHub page: \\ \url{https://github.com/scikit-learn-contrib/DESlib}. 

\end{abstract}


\section{Introduction}

Dynamic selection (DS) has become an active research topic in the multiple classifier systems literature in recent years. In this paradigm, one or more base classifiers\footnote{The term base classifier refers to a single classifier belonging to an ensemble or a pool of classifiers.} are selected for each query instance to be classified.  Such techniques have demonstrated improvements over traditional (static) combination approaches, such as majority voting and Boosting~\citep{CRUZ2018195}. DS techniques work by estimating the competence level of each classifier from a pool of classifiers. Only the most competent, or an ensemble containing the most competent classifiers is selected to predict the label of a specific test sample. The rationale for such techniques is that not every classifier in the pool is an expert in classifying all unknown samples; rather, each base classifier is an expert in a different local region of the feature space. 

In this paper, we introduce a library for dynamic ensembles in python: DESlib. The library contains the implementation of the key dynamic selection techniques in the literature. It also provides static ensemble methods which are often used as baseline comparisons for dynamic ensembles. The following sections present the project organization, the API design, currently implemented methods and future directions for the API.

\section{Project management}

DESlib was developed with two objectives in mind: to make it easy to integrate Dynamic Selection algorithms to machine learning projects, and to facilitate research on this topic, by providing implementations of the main DES and DCS methods, as well as the commonly used baseline methods. Each algorithm implements the main methods in the scikit-learn API~\citep{scikit-learn}: \textbf{fit}(X, y), \textbf{predict}(X), \textbf{predict\_proba}(X) and \textbf{score}(X, y). Any classifier from scikit-learn (or from other libraries that follow this API) can be used as base classifiers, making the library easy to use and to integrate in other projects. 

The implementation of the DS methods is modular, following a taxonomy defined in \cite{CRUZ2018195}. This taxonomy considers the main characteristics of DS methods, that are centered in three components: (1) the  methodology used to define the local region, in which the competence level of the base classifiers are estimated (region of competence); (2) the source of information used to estimate the competence and (3) the selection approach to define the best classifier (for DCS) or the best set of classifiers (for DES). This modular approach makes it easy for researchers to implement new DS methods, in many cases requiring only the implementation of methods \texttt{estimate\_competence} and \texttt{select}.

The library is written in pure python, working on any platform, and depends on the following python packages: scikit-learn, numpy and scipy. The project follows these guidelines:
\vspace{-0.3em}
\begin{itemize}[noitemsep]
	
	\item \textbf{Development:} All development is performed collaboratively using GitHub and Gitter, which facilitates code integration, communication between collaborators and issue tracking. External contributions are encouraged.
	
	\item  \textbf{Code quality:} The code was written following the PEP 8 standards. We use Codacy\footnote{https://codacy.com/} to measure and track code quality. The library is also covered by unit tests (py.test), using Travis CI. Moreover, Codacy and Travis CI are used to automatically check each new contribution according to the code quality and test coverage. 
	
	\item  \textbf{Documentation:} The code of DESlib is fully documented, including detailed instructions and examples  for using the API. The documentation is provided based using numpydoc and sphinx, being automatically updated with new developments. It is available online at \url{http://deslib.readthedocs.io/en/latest/}
	
	\item  \textbf{Bugs and new features:} Bugs and new feature requests are tracked through the project's GitHub page: \url{https://github.com/scikit-learn-contrib/DESlib/issues}. This environment allows a discussion between the collaborators to find the best solution for the problem. New users can check whether the problems they found or new requests are already being addressed.
	
	\item  \textbf{Project relevance:} At the edition time, the library is on its third release (v0.3), counts with 7 contributors (2 main and 5 external), and attracts about 500 new visitors weekly. Moreover, it is part of the scikit-learn-contrib supported projects.

\end{itemize}

\section{Implemented techniques}

The library is divided into three modules:
\vspace{-0.5em}
\begin{itemize}[noitemsep]
	\item \textbf{Dynamic Classifier Selection (DCS):} This module contains the implementation of techniques in which only the base classifier that attained the highest competence level is selected for the classification of the query.
	
	\item \textbf{Dynamic Ensemble Selection (DES):} Dynamic ensemble selection strategies refer to techniques that select an ensemble of classifier rather than a single one. All base classifiers that attain a minimum competence level are selected to compose the ensemble of classifiers.
	
	\item \textbf{Static Ensembles:} This module provides the implementation of static ensemble techniques that are usually used as a baseline for the comparison of DS methods: Single Best (SB), Static Selection (SS), Oracle and Stacked Generalization.
	
\end{itemize}

Tables \ref{tab:methods} and \ref{tab:baseline_methods} list the implemented DS and baseline methods, respectively.

\begin{table}[H]
\centering
	\caption{Implemented DES and DCS methods}
	\label{tab:methods}
	\resizebox{0.70\textwidth}{!}{  
	\begin{tabular}{ll}
		\toprule
		
		DES                                            & DCS                                      \\      
		
		\midrule
		
		META-DES \citep{CruzPR}    & Modified Rank \citep{classrank}             \\
		
		KNORA-E~\citep{knora}          			 & OLA \citep{lca}         						   \\
		
		KNORA-U \citep{knora}           		 & LCA \citep{lca}             				    \\
		
		DES-P \citep{WoloszynskiKPS12} 			 & MLA \citep{Smits_2002}                                  \\
		
		KNOP \citep{paulo2}             & MCB \citep{mcb}                          \\
		
		DES-RRC \citep{Woloszynski}     		 & A Priori \citep{DidaciGRM05}                       \\
		
		DES-KL \citep{WoloszynskiKPS12}   	 & A Posteriori \citep{DidaciGRM05}              \\
		
		DES-Exponential \citep{WoloszynskiK09}                 & Online Local Pool \citep{souza2019online}                                                   \\
		
		DES-Logarithmic \citep{WoloszynskiK09}                  &                                                   \\
		
		DES-Minimum Difference  \citep{zbMATH05935973}             &                                                \\
		
		DES-Clustering \citep{de2008empirical}                         &                                              \\
		
		DES-KNN \citep{de2008empirical}                           &                                       	\\
		
		DES-Multiclass Imbalance (DES-MI) \citep{garcia2018dynamic} 	&  \\
		
		\bottomrule
	\end{tabular}}
\end{table}

\begin{table}[H]
	\centering
	\caption{Implemented baseline methods}
		\resizebox{0.30\textwidth}{!}{  
	\label{tab:baseline_methods}
	\begin{tabular}{l}
		\toprule
		Static   \\      
		
		\midrule
		
		Oracle \citep{Kuncheva:2002}         \\
		
		Single Best     \\
		
		Static Selection \citep{classmaj}   \\
		
		Stacked Generalization \citep{wolpert1992stacked} \\
	
		\bottomrule
	\end{tabular}}
\end{table}

The library also provides several state-of-the-art improvements to DS techniques, such as the online Dynamic Frienemy Pruning (DFP) algorithm used in the FIRE-DES framework~\citep{Oliveira2017, cruz2019fire}, as well as 
dynamic weighting and hybrid selection + weighting~\cite{ijcnn2015} versions of DES techniques. 

\section{Installation and Usage}

The latest stable version of the library can be installed using pip (Python package manager): \texttt{pip install deslib}. Alternatively, the master branch, which contains features that will be included in future releases, can be installed directly from the GitHub address: \texttt{pip install git+https://github.com/scikit-learn-contrib/deslib}. New features are only merged to the master branch after code review and the creation of unit tests.

\subsection{Usage}

Each implemented method receives as an input a list of classifiers. This list can be either homogeneous (i.e., all base classifiers are of the same type) or heterogeneous (base classifiers of different types). The library supports any type of base classifiers from scikit-learn.

After instantiation, the method \textbf{fit}(X, y) is used to fit the Dynamic Selection method. Predictions for new examples can then be obtained with \textbf{predict}(X) and \textbf{predict\_proba}(X). In the example below, we show how to use the library, with a given Training (X\_train, y\_train), and Testing (X\_test, y\_test) datasets. The META-DES~\citep{CruzPR} technique is used in this example:

\begin{lstlisting}[language=Python]
from sklearn.ensemble import RandomForestClassifier
from deslib.des.meta_des import METADES

# Train a pool of 10 classifiers
pool_classifiers = RandomForestClassifier(n_estimators=10)
pool_classifiers.fit(X_train, y_train)
# Initialize the DS model
metades = METADES(pool_classifiers)

# Fit the dynamic selection model
metades.fit(X_dsel, y_dsel)

# Predict new examples:
metades.predict(X_test)
\end{lstlisting}

As of version 0.3, each implemented method comes with a list of default values, not requiring a trained list of classifiers as input. In such case, the pool of classifiers is trained together with the DS algorithm inside the \textbf{fit} method. More examples of using different aspects of the library can be found on \url{https://deslib.readthedocs.io/en/latest/auto_examples/index.html}.

\section{Conclusion and future plans}

In this paper, we introduced the DESlib, a Python library with the implementation of the state-of-the-art dynamic classifier and ensemble selection techniques. The project is fully compatible with the scikit-learn API and is part of the scikit-learn-contrib supported projects. Future work on this library includes the implementation of dynamic selection methods in different contexts, such as One-Class-Classification (OCC) and regression.

\bibliography{report}
\bibliographystyle{abbrv}

\end{document}